# Evaluating the Visual Similarity of Southwest China's Ethnic Minority Brocade Based on Deep Learning

Shichen Liu, Huaxing Lu*

School of Geomatics Science and Technology (Nanjing Tech University), Nanjing , China

**Abstract**:This paper employs deep learning methods to investigate the visual similarity of ethnic minority patterns in Southwest China. A customized SResNet-18 network was developed, achieving an accuracy of 98.7% on the test set, outperforming ResNet-18, VGGNet-16, and AlexNet. The extracted feature vectors from SResNet-18 were evaluated using three metrics: cosine similarity, Euclidean distance, and Manhattan distance. The analysis results were visually represented on an ethnic thematic map, highlighting the connections between ethnic patterns and their regional distributions.

## 1 Introduction

Ethnic patterns play a crucial role in various cultures, reflecting the unique characteristics and identities of different groups through their distinctive designs and symbols. Previous research has explored this topic from multiple perspectives. For example, Wang [1] examined the aesthetic values, emotional significance, and taboos embedded in ethnic patterns. Liu [2] emphasized the close relationship between these patterns and the rich history and culture of the Chinese nation, while Wang Hua et al. [3] studied the impact of regional differences on ethnic patterns and styles. Understanding the similarities and regional distribution of ethnic patterns offers deeper insights into the cultural connections among different ethnic groups. However, traditional methods for studying these patterns have primarily relied on manual analysis, which is not only inefficient but also subject to bias, making it difficult to apply in large-scale and complex data analysis scenarios. In recent years, the rapid advancement of deep learning techniques has provided new tools to address these challenges, particularly in image feature extraction. For instance, Lao et al. [4] developed a network for clothing feature extraction, utilizing the K-nearest neighbors algorithm to classify and retrieve clothing types and attributes. Dong et al. [5] integrated a spatial pyramid pooling mechanism with the VGGNet model[6], expanding the capacity for clothing image feature extraction and improving recognition accuracy. However, compared to the general field of clothing image analysis, research focused on the specific subfield of ethnic costume patterns remains limited. To address this gap and further advance the field, this paper introduces a deep learning-based similarity evaluation method using a convolutional neural network (SResNet-18). This method employs SResNet-18 to extract feature vectors from different ethnic groups and measure their similarity. Additionally, the study visualized the results on ethnic thematic maps, explored the relationship between ethnic patterns and their regional distribution, and further uncovered the underlying patterns of ethnic cultural diversity and distribution.

## 2 Method

### 2.1 Improved ResNET18 Network Model

ResNet is a classic deep convolutional neural network introduced by He et al.[7] in 2015. Its primary advantage lies in the ability to increase network depth, thereby capturing richer and more complex features, while addressing issues related to vanishing gradients and training difficulties. The core innovation of ResNet is the introduction of residual blocks, as illustrated in Figure 1.The input to the residual block can bypass intermediate layers and be directly passed to subsequent layers through cross-layer connections, which facilitates easier training of the network. These cross-layer connections help prevent the vanishing gradient problem that commonly occurs in traditional networks during backpropagation. The output of the residual block can be expressed by the following formula:
$$y = F(X) + x$$
Where $x$ represents the residual function, $F(X)$ and $y$ represents the input and output, respectively. Compared with other models, ResNet can extract image features [8,9], so ResNet was used as the main structure for feature extraction in this study.

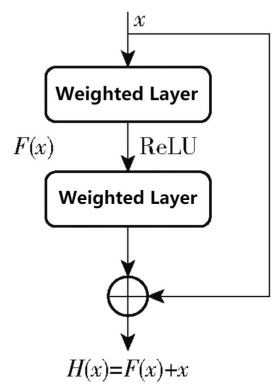

Figure 1: Residual block structure diagram.

While ResNet excels in various image recognition and classification tasks, it faces certain limitations in feature extraction when dealing with complex ethnic patterns. To overcome these challenges, this study incorporates the SE module[10] introduced by Jie Hu et al. in 2018. The SE module enhances network performance by optimizing inter-channel dependencies through a self-attention mechanism.The SE module operates through two primary steps: Squeeze and Excitation. During the Squeeze phase, global average pooling is applied to condense the information from each feature channel, generating a vector that represents the importance of each channel. In the subsequent Excitation phase, this vector is input into a fully connected network to calculate adaptive weights for each channel.

In this study, the SE module was embedded between the Conv3 and Conv4 layers of ResNet-18, resulting in a new network architecture named SResNet-18 (as shown in Figure 2). In this architecture, the first two convolutional layers, Conv1 and Conv2, perform the initial feature extraction. The SE module is then applied to the output feature map of Conv3 for fine-grained channel reweighting. Finally, the weighted feature map is passed to the Conv4 layer, allowing the subsequent layers to continue processing. These enhancements enable the network to learn and represent useful features more effectively, thereby improving performance and enhancing the extraction of ethnic pattern features.

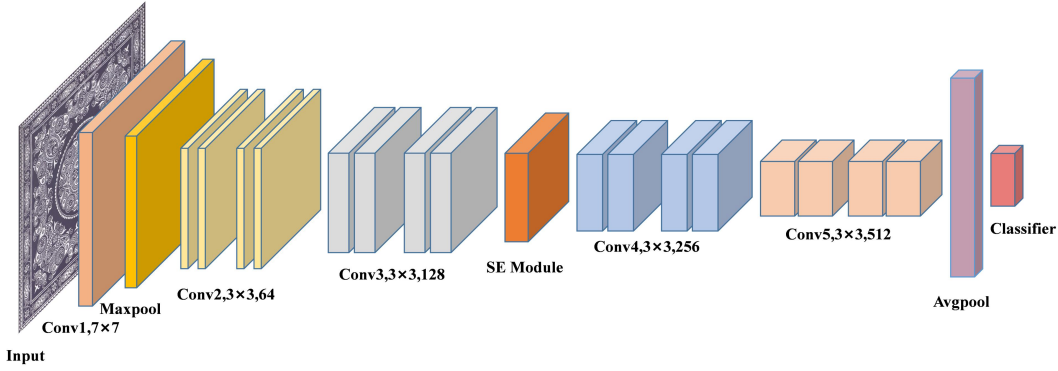

Figure 2. ResNet-18 network structure diagram.

## 2.2 Feature Vector Extraction

Leveraging the feature extraction capabilities of convolutional neural networks, the model $f_\theta$ generates a corresponding feature vector $z = f_\theta(x)$ for a given input image x. When the model is set to inference mode, the weight parameters remain fixed and are not updated during forward propagation. Through this process, the model produces a feature vector that encapsulates the critical features learned from the input image.

Using the feature extraction ability of the convolutional neural network, for the given input image $x$, the corresponding feature vector $z = f_\theta(x)$ can be obtained through the model $f_\theta$. When the model is set to inference mode, the weight parameters remain fixed and are not updated during forward propagation. Through forward propagation, the model generates a feature vector based on the extracted features of the input image. This feature vector captures the important features that the image learns in the model.

Inspired by the idea of global average pooling (GAP) [11], the feature vectors of all images $z$ in the same category are averaged to obtain a representative feature vector. This representative eigenvector $v_i$ can be calculated by the following mathematical formula:

$$v_i = \frac{1}{N_i} \sum_{j=1}^{N_i} z_j^i$$

Where, $N_i$ represents the number of images in the category $i$, $z_j^i$ represents the feature vector of the $i$ image in the category $j$. By averaging the feature vectors, we reduce the influence of possible noise or outliers on the feature vectors in a single image, thus improving the stability and representation of the feature vectors. This approach not only enhances the robustness of feature vectors to changes within categories, but also retains the expressive power of uniqueness between categories.

## 2.3 Cosine Similarity

Cosine similarity is a quantitative method used to evaluate vector similarity by calculating the cosine of the angle between two vectors in an inner product space. In digital image processing, cosine similarity is particularly effective for quantifying similarity between images, as feature vectors typically reside in high-dimensional spaces. Each ethnic pattern image feature can be represented as a vector in a high-dimensional space, and their similarity is evaluated by calculating the cosine of the angle between these vectors. When two vectors point in nearly the same direction in this high-dimensional space, their cosine similarity approaches 1, indicating a high degree of similarity. Conversely, if the vectors differ significantly in direction, the cosine similarity approaches -1, indicating low similarity. The formula for calculating the cosine similarity between two vectors $X$ and $Y$ in $N$ dimensions is:

$$\cos(\theta) = \frac{XY}{||A||\,||B||} \frac{\sum_{i=1}^{n} X_i \times Y_i}{\sum_{i=1}^{n}(X_i)^2 \times \sum_{i=1}^{n}(Y_i)^2}$$

## 3 Experiments
### 3.1 Dataset
This study selected ethnic brocade patterns from Southwest China as experimental data, primarily sourced from museum costume collections and field visits. The dataset comprises images from 10 different ethnic groups, including Dong, Yao, Miao, Qiang, Shui, Tujia, Yi, Zhuang, Bai, and Li, with a total of 1,430 images (143 images per ethnic group), as shown in Figure 3. A total of 3,330 images were selected for ethnic pattern analysis, with the remaining 1,100 images split in a 7:3 ratio, 770 images were allocated for training and 330 for testing. Given the limited size of the training set, data augmentation techniques such as rotation, flipping, zooming, translation, and resizing were applied to expand the dataset.

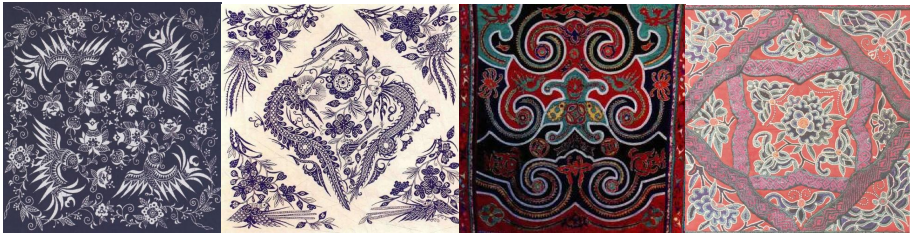

Figure 3. Visual representation of ethnic patterns.

### 3.2 Loss Function and Optimizer
The loss function measures the discrepancy between the model's predictions and the true values. Typically, a smaller loss function value indicates that the model's predictions are closer to the actual values[12]. The model parameters that minimize the loss function are considered the optimal parameters. In this study, the cross-entropy loss function was selected to evaluate the alignment between the model's predictions and the target categories, using the following formula:

$$loss(x, class) = -\log\left(\frac{\exp(x[class])}{\sum_j \exp(x[j])}\right) = -x[class] + \log\left(\sum_j \exp(x[j])\right)$$

Where x is the result of the model generation, and class is the label corresponding to the data.

Adam (Adaptive Moment Estimation) is an optimization algorithm commonly used in deep learning models as an alternative to stochastic gradient descent. This algorithm combines two gradient descent variants (RMSProp and momentum gradient descent) to maintain fast training speeds while also improving the likelihood of finding a global optimal solution in the parameter space[13] . Consequently, the Adam optimizer was employed in the model training for this study.

### 3.3 Model Training
The experimental environment for this study consisted of an AMD R7-5800H processor, 16GB of memory, and an NVIDIA GeForce RTX 3060 GPU, all running on a Windows operating system. The deep learning training platform was built using the PyTorch 1.12 framework with Python 3.9 and CUDA version 11.3.The performance of the convolutional neural network (CNN) model is highly dependent on the model parameter settings. In this study, a batch size of 64 was used, and the model was trained for 50 epochs with an initial learning rate of 0.001. Before training, the input images were preprocessed and standardized to enhance the model's feature extraction capabilities and ensure the input data was within an optimal range for training.Given the small size of the ethnicity pattern dataset, transfer learning[14] was employed. The SResNet-18 model was pre-trained on a large dataset from ImageNet, and the resulting weights were then fine-tuned on the pattern dataset.

## 4. Results and Discussion

### 4.1 Model Training Results

In this experiment, the SResNet-18 model was trained for 50 cycles using transfer learning. The model's performance was assessed by tracking both the loss value and accuracy across each training epoch. As shown in Figure 4, the SResNet-18 model achieved 98.7% accuracy on the training set, compared to 97.27%, 97.01%, and 96.23% for the original ResNet-18, VGGNet-16, and AlexNet[15], respectively. These comparative experiments demonstrate that embedding the SE module provides SResNet-18 with a clear advantage in feature extraction capability and identification accuracy over the other models.

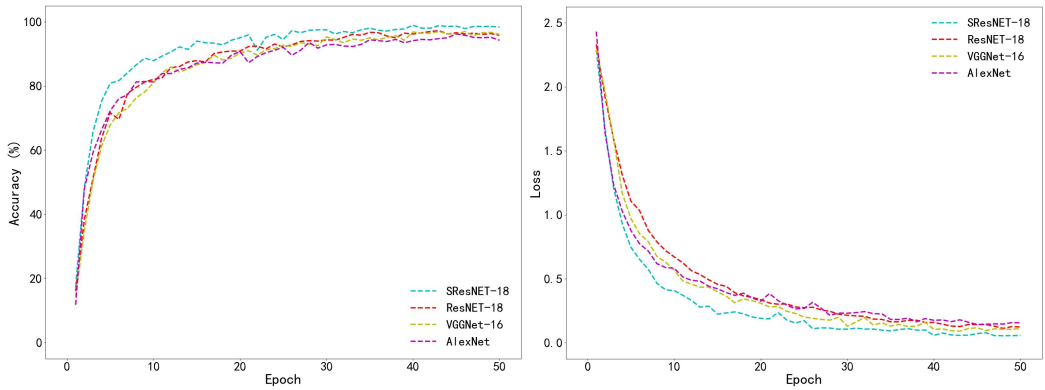

**Figure 4. Accuracy and loss plots across training epochs.**

### 4.2 Ablation Experiments

To evaluate the impact of incorporating attention mechanisms into the network, this study conducted ablation experiments comparing the effects of attention mechanisms applied at different stages of the ResNet-18 architecture. The four configurations tested were: attention mechanism before Conv2 (Scheme 1), between Conv2 and Conv3 (Scheme 2), between Conv3 and Conv4 (Scheme 3), and between Conv4 and Conv5 (Scheme 4). Using the same dataset and training parameters, the experimental results are presented in Table 1.

**Table 1. Performance of 4 schemes**

| Scheme | Scheme 1 | Scheme 2 | Scheme 3 | Scheme 4 |
|---|---|---|---|---|
| Precision /% | 98.1 | 98.4 | 98.7 | 98.5 |

As indicated in Table 1, the model's accuracy improves when the attention mechanism is applied at deeper layers within the network, particularly between Conv3 and Conv4. However, given the relatively small dataset, introducing the attention mechanism at even deeper layers, such as between Conv4 and Conv5, increases the risk of overfitting, leading to a reduction in accuracy.

### 4.3 Visualization of the Measurement Results

To quantitatively analyze the differences among ethnic minority patterns in China, cosine similarity, Euclidean distance, and Manhattan distance were employed to measure the features extracted by the SResNet-18 model. The central element of the Shui ethnic group's pattern was used as the reference point on the ethnic vector map, and the distances between all ethnic patterns and the Shui patterns were calculated. This approach aimed to more accurately capture the cultural differences between the various ethnic patterns. The results of the three measurement methods are presented in Table 2.

Table 2. Ethnic pattern feature vector distance under different measures

| ethnic categories | Euclidean distance | Manhattan distance | cosine similarity |
|---|---|---|---|
| Dong | 18.2790 | 40.8378 | 0.0416 |
| Tujia | 16.9534 | 34.2012 | 0.0062 |
| Zhuang | 12.4965 | 28.7272 | 0.3391 |
| Yi | 13.6298 | 29.8265 | 0.2468 |
| Shui | 0 | 0 | 1 |
| Yao | 17.8603 | 32.0379 | 0.0437 |
| Bai | 19.0577 | 38.7833 | -0.4534 |
| Qiang | 19.0773 | 38.4020 | -0.5184 |
| Miao | 17.6651 | 38.3935 | 0.1960 |
| Li | 13.8291 | 28.9968 | 0.2886 |

By visualizing the measurement results on an ethnic thematic map, the differential distribution of Southwest China's ethnic minority patterns is intuitively displayed (Figure 5). Each ethnic group is represented by a vector on the map, with color variations highlighting the similarities and differences between the patterns of different ethnic groups. The comparison of the three measurement methods on the thematic map revealed that cosine similarity effectively highlights the similarities and differences among ethnic patterns. Unlike Euclidean distance and Manhattan distance, cosine similarity emphasizes the angle between two feature vectors rather than their absolute magnitude. This characteristic allows cosine similarity to more accurately capture the nuances of similarity or difference in high-dimensional feature spaces. The visual representation of ethnic patterns using this method offers a comprehensive and intuitive approach to understanding the relationships between different ethnic patterns.

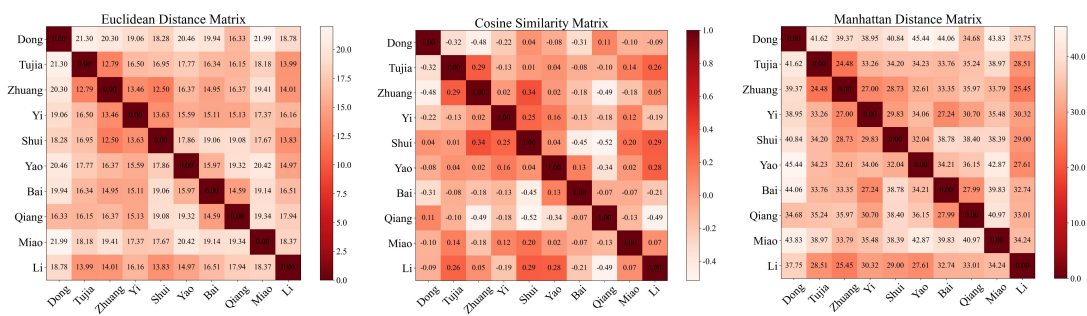

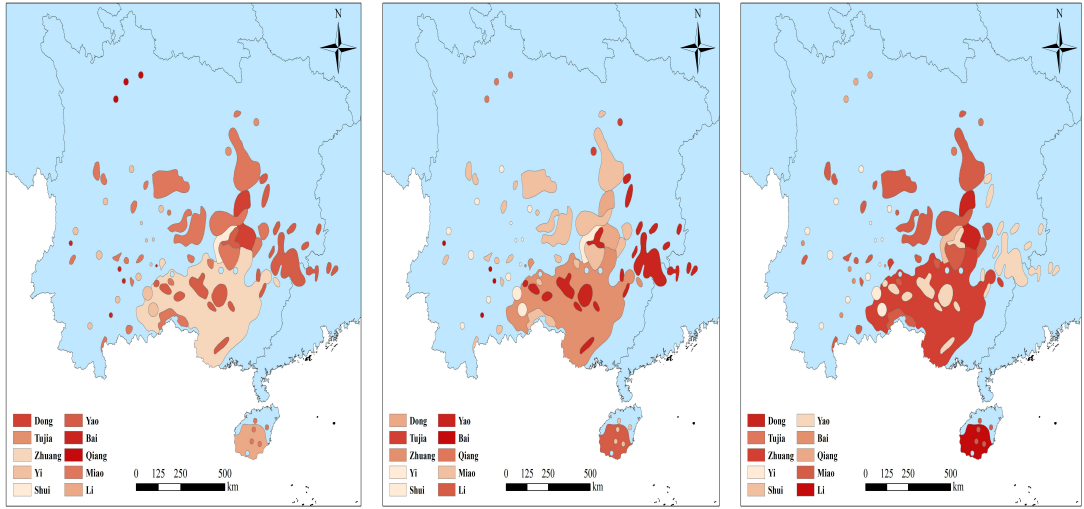

Figure 5. Results of different measures

## 5. Conclusions

    Ethnic patterns are not only a form of artistic expression but also serve as significant carriers of national cultural identity. To analyze the visual similarity of ethnic minority patterns in China, this paper proposes a similarity evaluation method based on deep learning. The SE module was integrated into ResNet-18, resulting in a new network architecture named SResNet-18. This enhanced architecture significantly improved the feature extraction capabilities and recognition accuracy due to the SE module's contribution. Experimental results demonstrated that SResNet-18 achieved a 98.7% accuracy on the test set, slightly outperforming ResNet-18, VGGNet-16, and AlexNet. This confirms SResNet-18's superior ability in extracting features from ethnic pattern images. For the analysis, feature vectors were extracted using SResNet-18 and evaluated with three different measures: cosine similarity, Euclidean distance, and Manhattan distance. The results of these measurements were visualized on a national thematic map, where the cosine similarity method proved particularly effective in revealing the similarities and differences between ethnic patterns. Visualizing the measurement results of ethnic patterns offers a comprehensive and intuitive approach to understanding the relationships between different ethnic groups. Additionally, it highlights the association between cultural patterns and their geographical distribution, providing an exemplary analytical method for future research in ethnic studies.